\documentclass[conference]{IEEEtran}
\IEEEoverridecommandlockouts
\usepackage[utf8]{inputenc}
\usepackage[compress]{cite}
\usepackage{amsmath,amssymb,amsfonts}
\usepackage{algorithmic}
\usepackage{graphicx}
\usepackage{textcomp}
\usepackage{xcolor}
\usepackage{subfigure}
\usepackage{tikz} 

\graphicspath{{img/}}
\title{
Understanding Spending Behavior: Recurrent Neural Network Explanation and Interpretation
\thanks{This work is partially funded by The Norwegian Research Foundation, project number 311465.}
}

\author{\IEEEauthorblockN{Charl Maree\IEEEauthorrefmark{1}}
\IEEEauthorblockA{
\textit{Center for Artificial Intelligence Research} \\
\textit{University of Agder}\\
Grimstad, Norway \\
charl.maree@uia.no} 
\thanks{\IEEEauthorrefmark{1}Author's second affiliation: Technology, Development, and Business Processes, Sparebank 1 SR-Bank, Stavanger, Norway.}
\and
\IEEEauthorblockN{Christian W. Omlin}
\IEEEauthorblockA{
\textit{Center for Artificial Intelligence Research} \\
\textit{University of Agder}\\
Grimstad, Norway \\
christian.omlin@uia.no}
}
\begin{document}

\maketitle
\begin{abstract} 
Micro-segmentation of customers in the finance sector is a non-trivial task and has been an atypical omission from recent scientific literature. Where traditional segmentation classifies customers based on coarse features such as demographics, micro-segmentation depicts more nuanced differences between individuals, bringing forth several advantages including the potential for improved personalization in financial services. AI and representation learning offer a unique opportunity to solve the problem of micro-segmentation. Although ubiquitous in many industries, the proliferation of AI in sensitive industries such as finance has become contingent on the explainability of deep models. We had previously solved the micro-segmentation problem by extracting temporal features from the state space of a recurrent neural network (RNN). However, due to the inherent opacity of RNNs our solution lacked an explanation. In this study, we address this issue by extracting a symbolic explanation for our model and providing an interpretation of our temporal features. For the explanation, we use a linear regression model to reconstruct the features in the state space with high fidelity. We show that our linear regression coefficients have not only learned the rules used to recreate the features, but have also learned the relationships that were not directly evident in the raw data. Finally, we propose a novel method to interpret the dynamics of the state space by using the principles of inverse regression and dynamical systems to locate and label a set of attractors. \\
\end{abstract}

\begin{IEEEkeywords}
Explainable AI, Micro-segmentation, Inverse regression, Dynamical systems
\end{IEEEkeywords}
\section{Introduction} 
Customer segmentation is an important field in banking and with customer bases growing banks are having to employ ever advancing methods to maintain, if not improve, levels of personalization \cite{Matteo19}. Customer segmentation has typically been achieved using demographics such as age, gender, location, etc. \cite{Kalia18}. However, these features not only produce coarse segments, but also introduce the potential for discrimination, e.g., when using postal codes for credit rating \cite{Barocas16}. In contrast, micro-segmentation provides a more sophisticated, fine-grained classification that depicts nuanced differences between individuals, improves personalization, and promotes fairness. Despite these advantages and the fact that the need for such fine-grained segmentation has been highlighted \cite{Krishnapuram17}, the scientific community has been surprisingly quiet on the topic with only a few recent publications from, e.g., the health sector \cite{Kuwayama19, Nandapala20} and apparently none from the finance sector.

Artificial intelligence is fast becoming ubiquitous across multiple industries with representation learning an auspicious method for customer micro-segmentation \cite{maree21}. Sensitive industries such as finance face legal and ethical obligations towards the responsible implementation of AI \cite{vdBurgt19}. The European Commission has published several guidelines surrounding responsible AI and scientific fundamentals have been consolidated in recent surveys on the topic \cite{EU20, Arrierta20}. Explainability and interpretability are key elements in responsible AI \cite{Goodman17}, which are generally not yet adequately addressed in applications of AI in finance \cite{Cao21}. Our perspective on explainability in AI refers to a symbolic representation of a model, whereas interpretability refers to a human understanding of and reasoning about the functionality of the model. Explainability therefore neither guarantees nor implies interpretability. In this study, we address both the issues of explainability and interpretability and we introduce a novel method for interpretation of features based on inverse regression and dynamical systems \cite{parker, ceni19}.

Our aim is to extract and facilitate the use of salient features in future financial services; we have already shown the potential in predicting default rate and customer liquidity indices \cite{maree21}. Our ultimate goal is the development of personalized financial services in which responsible customer micro-segmentation is key.
\section{Related Work} 
\label{rel}
\subsection{Representation Learning using Recurrent Neural Networks} 
In \cite{glad16}, the authors developed a model for predicting spending personality from aggregated financial transactions with the intent to investigate the causality between personality-aligned spending and happiness. They rated each of 59 spending categories according to its association with the Big-Five personality traits - extraversion, neuroticism, openness, conscientiousness, and agreeableness \cite{DeRaad00} - which resulted in a set of $59 \times 5$ linear coefficients. We used these coefficients in a previous study to train a recurrent neural network (RNN) to predict customers' personality traits from their aggregated transactions \cite{maree21}. We showed that the temporal features in the state space of the RNN had interesting properties: they formed smooth trajectories which formed hierarchical clusters along successive levels of dominance of the personality traits. The dominant personalty trait is the one with the largest coefficient in the Big-Five model of personality traits \cite{maree21}. We also showed that similarly salient features could not be extracted from the raw data otherwise. Spending patterns over time are either more consistent than transactions aggregated over a short time period, they may fluctuate, or they may change based on life circumstances. Modelling spending over time reinforces spending patterns and thus may lead to better features. Fluctuations or changes are also better represented by time series. The hierarchical clustering of the extracted features provided a means of micro-segmenting customers based on their financial behaviour. However, the responsible employment of this model demands an explanation and interpretation, which is what we address in this study. Other studies have employed representation learning using RNNs to encode spatial and temporal information contained in the two-dimensional trajectories of physical objects \cite{yao17}, as well as for a wide variety of applications in speech and video data \cite{Li20}.

\subsection{Explaining Recurrent Neural Networks}
Finding symbolic representations of AI models is a key area of explainable AI \cite{Arrierta20}. In \cite{udrescu20} the authors developed a symbolic regression algorithm that successfully extracted physics equations from neural networks. They managed to extract all 100 of the equations from the well known Feynman Lectures on Physics and 90\% of more complicated equations, an improvement from 15\% using state-of-the-art software. This was an important study because it not only proved that deep neural networks are capable of learning complicated equations and coefficients, but that it is possible to extract symbolic knowledge from such networks. The authors in \cite{ming17} presented a visual method to explain RNNs used in natural language processing problems. They clustered the activations in the state space and used word clouds to visualize correlations between node activations and words in the input sentences. Similarly, the authors in \cite{omlin96} applied clustering in the state space of RNNs, but here the authors showed that \emph{symbolic} representations could be extracted as opposed to visual explanations. Studies such as these prove that deep neural networks are indeed not inexplicable black box systems, but could be a means of discovering symbolic representations of complex relationships in data.
\section{Methodology} 
\label{exp}
\subsection{Recurrent Neural Network Training}
We used the financial transactions of approximately 26,000 customers to train a RNN to predict spending personality, as described in detail in \cite{maree21}. To summarize, the input data were each customer's transactions aggregated annually across 97 transaction classes, such as groceries, transport, leisure, etc., over a period of six years. This gave an input vector $I \in [0,1]^{N \times T \times C}$ where  $\sum_{c=1}^{C}{I_{n,t,c} = 1 \, \forall \, n\in [1,N], t \in [1,T]}$ where $N\simeq 26000$ customers, $T=6$ time-steps, and $C=97$ transaction classes. Each value in $I$ therefore represented the fraction of total income spent by a given customer in a given year on a given transaction class. The output data $O \in [-1,1]^{N \times P}$ were the customers' Big-Five personality traits (i.e. $P=5$) calculated from published linear coefficients linking transaction classes to personality traits \cite{glad16}. Our RNN consisted of three long short-term memory (LSTM) nodes \cite{lstm97}. The number of nodes was determined by optimizing the diminishingly increasing prediction accuracy for an increasing number of nodes, also known as the `elbow' optimization method; RNN architectures are known to perform well with low-dimensional representations \cite{mahes19}. After training and during prediction, we inspected the activations of the three recurrent nodes in the state space $S \in \mathbb{R}^{N \times T \times M}$ where $M=3$ is the number of LSTM nodes; each customer was represented by a trajectory with six data points in the three-dimensional space. These trajectories were our extracted features which may be used for micro-segmentation of customers \cite{maree21}. 

\subsection{Explanation through Surrogate Modelling}
To provide an explanation for the RNN, we trained a linear regression model - an inherently transparent class of models \cite{Arrierta20} - to replicate the trajectories from each customer's aggregated spending distribution: $F_{\theta}(I) \mapsto S$ where $\theta$ represents the coefficients of the linear regression model $F$. We show that these coefficients reproduced, with high-fidelity, the states of the RNN, thereby offering a symbolic explanation of its functioning.

\subsection{Interpretation through Inverse Regression}
\label{sec:inv_pred}
To obtain an interpretation of the features, we propose a new method that maps the output space $O$ onto the state space $S$ using inverse regression \cite{parker}. From an $M$-dimensional grid $S' \in \mathbb{R}^{|K| \times M}$ where $S_i' \in \{0.1k, k \in K=[-10,10]\}, i \in [1,..,M]$, filling the entire volume of the $M$-dimensional state space $S$, and using the trained weights of the \emph{output layer} of the RNN, $\omega_{out} \in \mathbb{R}^{M\times(P+1)}$ 
\footnote{The dimensions $M \times (P+1)$ represent the weights connecting the $M$ LSTM nodes to the $P$ output nodes, plus one dimension to account for the bias.}, 
we calculated the entire \emph{reachable} output as a $P$-dimensional hypercube $O' \in \mathbb{R}^{|K|\times P}$, where $|K|=21$ is the number of points in each dimension of the grid $S'$. Formally,

\begin{equation*}
O' = S' \cdot \omega_{out} 
\end{equation*}

\begin{figure}[!htbp] 
 \centering 
 \includegraphics[width=0.9\linewidth]{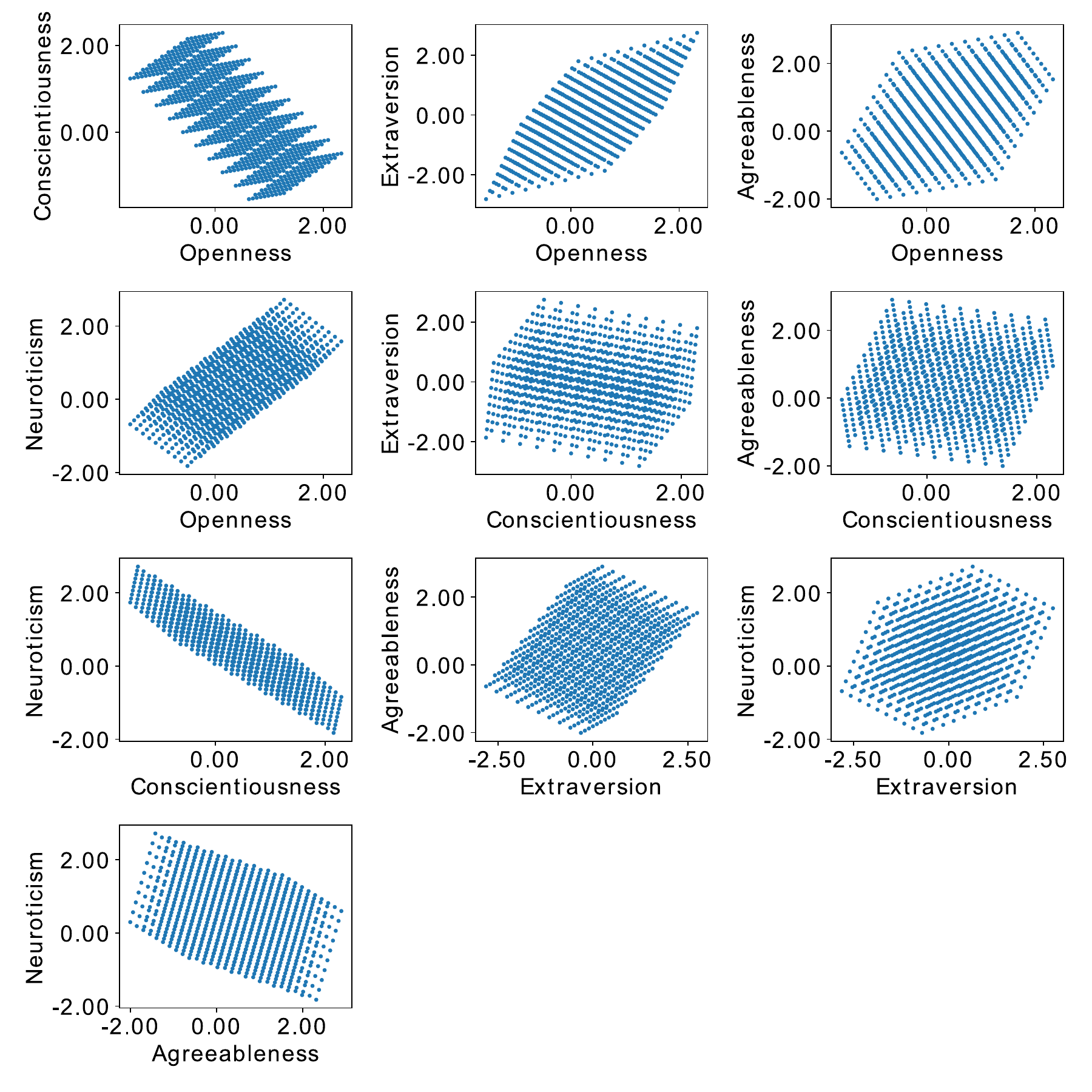} 
 \caption{The reachable output space of our RNN shown as two-dimensional projections of all combinations of the five output dimensions. The reachable output space was mapped from the reachable region in state space ($S' \in [-1..1]^3$) using the output weights of the RNN
 } 
 \label{fig:output_space} 
\end{figure}

This reachable hypercube of the output space is shown in Figure \ref{fig:output_space}. Next, using the principles of inverse regression as described in \cite{parker}, we calculated the parameters $\omega_{inv} \in \mathbb{R}^{(P+1)\times M}$ that map the output space $O$ to the state space $S$. Formally,

\begin{equation*}
\omega_{inv} = (O'^TO')^{-1} \cdot (O'^TS')
\end{equation*}

In order to map the \emph{magnitudes} of the dimensions of the output space $O$ onto the state space $S$, we created a diagonal matrix  $\mathcal{D} \in \mathbb{R}^{P \times P}$ with the elements on the diagonal equal to the magnitude of each dimension of the output hypercube $O'$: 

\begin{equation} \label{eqn:D}
\mathcal{D} = diag \left\{ \max_{1 \leq i \leq |K|}{O'_{i,j}, \ j \in [1..P]} \right\}
\end{equation}

The representation of the dimensions of the output space in the state space $\mathcal{O} \in \mathbb{R}^{P \times M}$ is then given by:

\begin{equation} \label{eqn:mapped_dim}
\mathcal{O} = \mathcal{D} \cdot \omega_{inv} - \bf{0}^P \cdot \omega_{inv}
\end{equation}

where $\bf{0}^P$ is the zero vector of size $P$ representing the origin of the output space and $\bf{0}^P \cdot \omega_{inv}$ is the location of this origin in the state space.
\section{Results}
\label{res}
\begin{figure}[!ht] 
 \centering
 \subfigure[]{
   \includegraphics[width=0.66\linewidth]{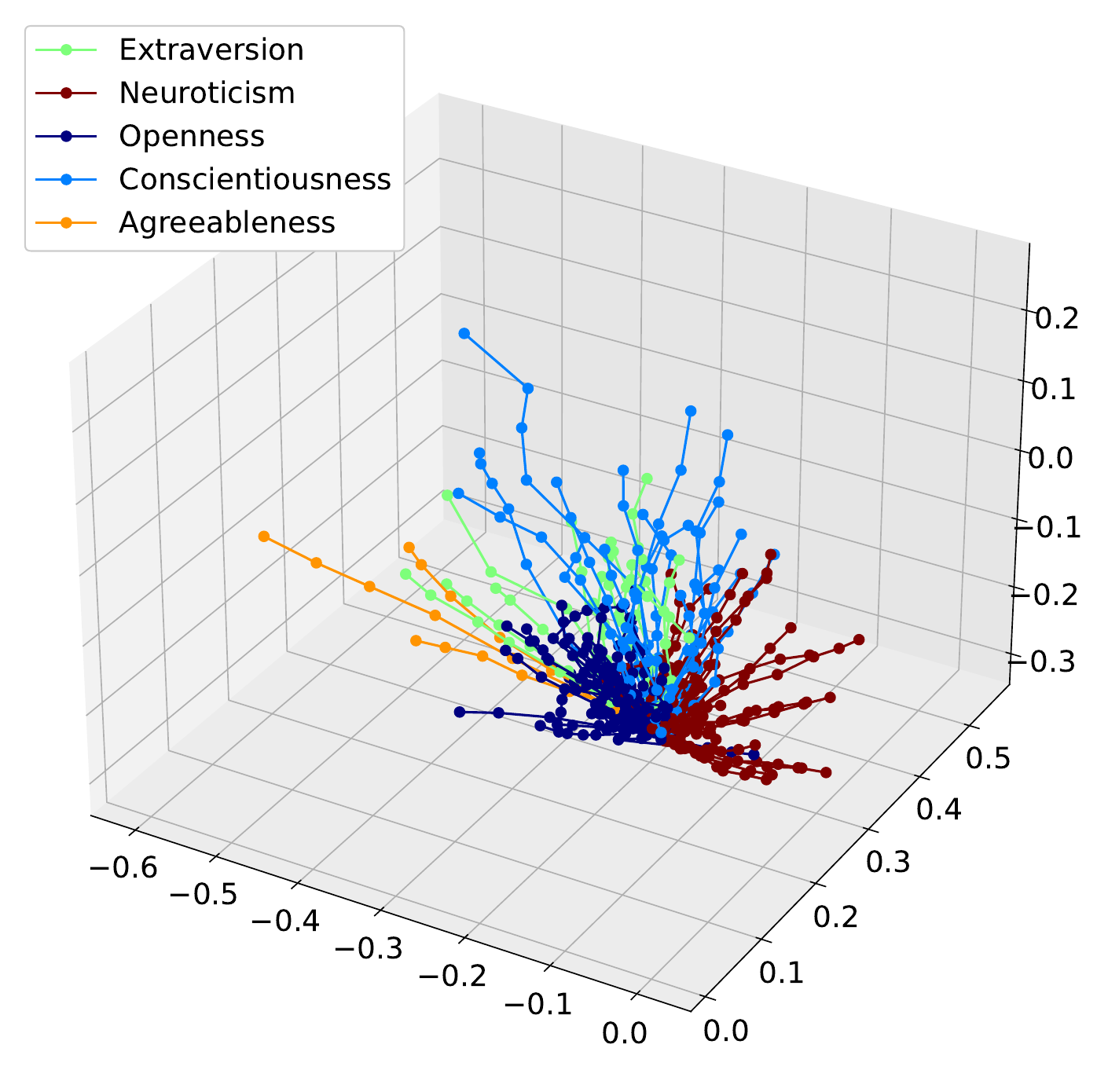}
   \label{fig:trajectories_a}
 }
 \\
 \subfigure[]{
   \includegraphics[width=0.66\linewidth]{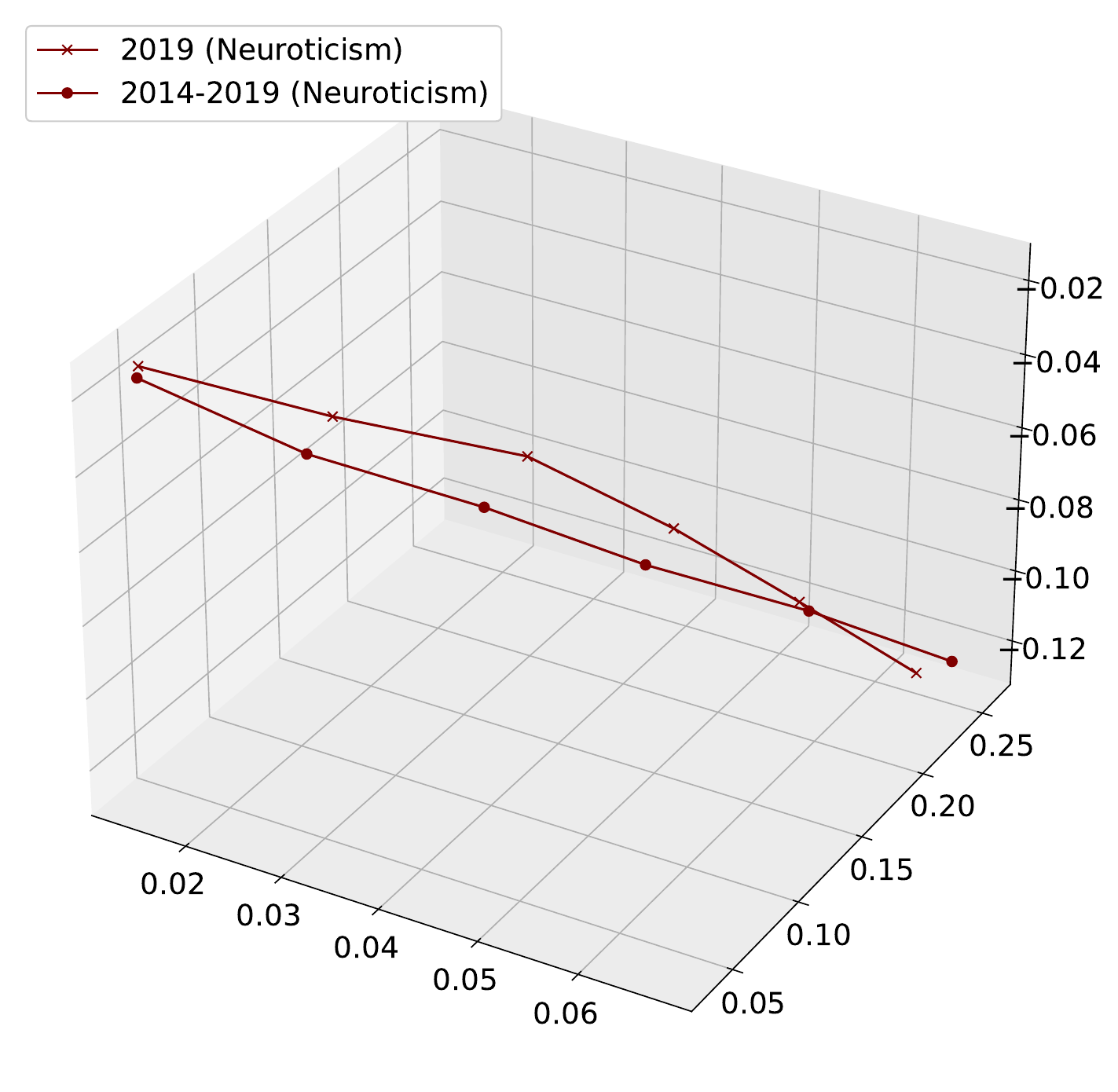}
   \label{fig:trajectories_b}
 }
 \\
 \subfigure[]{
   \includegraphics[width=0.66\linewidth]{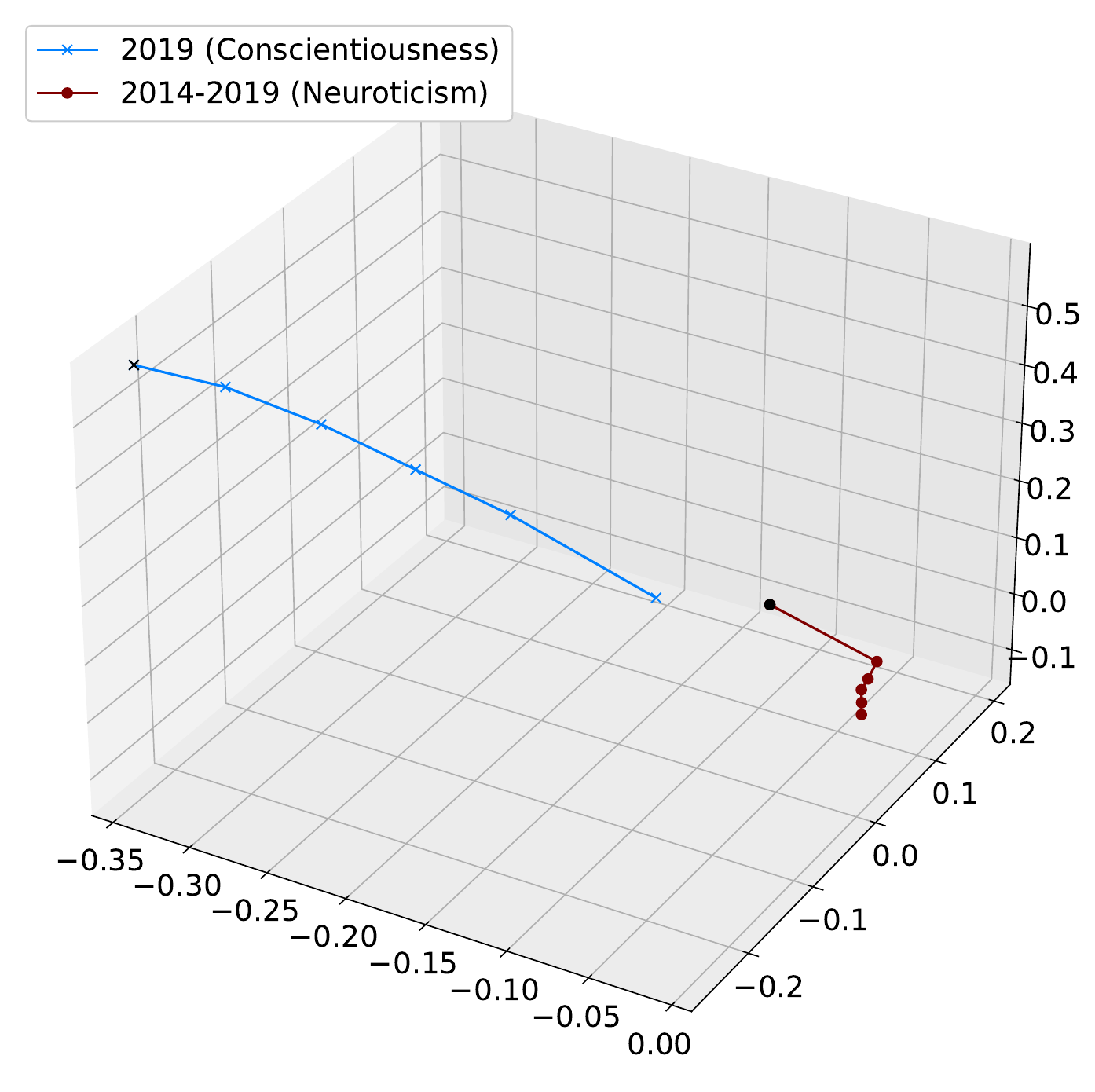}
   \label{fig:trajectories_c}
 }
 \caption{
  Trajectories in the 3-dimensional state space of a recurrent neural network trained to predict personality from aggregated transactions. While (a) shows the clustering of the trajectories of many customers according to their most dominant personality traits, (b) shows two trajectories for the same customer identically classified for two different time periods: one year vs. six years., and (c) again shows two such trajectories, but for a different customer that converged to a common attractor (conscientiousness) in the last year, after having converged to a different attractor (neuroticism) for the first five years.} 
 \label{fig:trajectories} 
\end{figure}

In Fig. \ref{fig:trajectories}, we show the features that we extracted from our RNN. Fig. \ref{fig:trajectories_a} illustrates the clustering behaviour of the trajectories in the state space. Our empirical observations led us to hypothesise the existence of attractors for each of the five personality traits. Fig. \ref{fig:trajectories_b} shows two trajectories for the same customer where the inputs to the RNN were aggregated over two different time periods: one year and six years. The fact that there is little difference between these two trajectories is significant; it demonstrates that the duration of the time window did not affect customer classification. This was not the case when clustering the raw personality data, where customers frequently moved between different clusters for different time periods due to variations in spending with changing life circumstances. Although we did observe significant course changes for some customers' trajectories (e.g., in Fig. \ref{fig:trajectories_c}), the vast majority of customers remained in their assigned clusters for the six-year period. This stability in customer micro-segmentation is key for personalized financial services, as financial advice has to be consistent. Fig. \ref{fig:trajectories_c} shows the long-term (six years) and short-term (one year) trajectories of a single customer who changed their spending behaviour such that their dominant personality type changed in the last year. In this figure it is clear that, for the final year, both trajectories moved towards the same attractor (conscientiousness), with the neuroticism attractor no longer acting upon the long-term trajectory. 

To explain our model, we fit a linear regression model to reproduce the trajectories in the state space $S$ from the RNN's input data $I$. From our observations in Fig. \ref{fig:trajectories_b}, we hypothesized that the lengths of the trajectories were not as important as their directions. We therefore simplified the trajectories and represented them by the two angles which fully describe their directions in three-dimensional space. These angles were the outputs of our linear regression model $F_{\theta}(I)$, which fit the data with a coefficient of determination of 0.78 for an unseen test set, while a more complicated polynomial regression model managed an only slightly better 0.79. Other methods such as ridge regression and decision tree regression were inferior in accuracy. Our 97 transaction classes mostly overlapped with those of the $59 \times 5$ published coefficients and due to aggregations such as "health and fitness" being expanded to "health" and "fitness", there were $61 \times 5$ non-zero coefficients for calculating our customers' personality traits. The linear regression model had $69 \times 2$ non-zero \footnote{Non-zero here refers to coefficients with values that are not insignificantly small compared to the mean value of all the coefficients.} coefficients with a strong correlation to the original non-zero coefficients. Further, within each of the clusters in Fig. \ref{fig:trajectories_a}, we observed hierarchical sub-clusters along the second, third, and fourth most dominant personality traits. This hierarchical sub-clustering is important because it provides a means of micro-segmenting customers which was not present in the raw data and could neither be replicated using feed-forward neural networks nor auto-encoders. Using our linear regression model, we created a two-dimensional plot of trajectory angles (Fig. \ref{fig:hierarchy}). In this figure, we illustrate the hierarchical clustering behaviour that we observed for the trajectories from the RNN, where (a) shows the clustering along the customers' most dominant personality trait and (b) through (d) show the hierarchy of sub-clusters within the parent clusters. These clusters, like the trajectory clusters, were consistent in time, i.e., the linear regression model retained the desirable properties of the features from the state space of our RNN. Due to this and the high accuracy obtained in testing, we conclude that the linear regression model matched the RNN with high fidelity. The parameters $\theta$ of the linear regression model are the symbolic explanation of the RNN, answering questions such as ``Why was Customer A classified in this way?'' by referring to the customer's aggregated transactions in the input data $I$.

\tikzset{boximg/.style={remember picture,thick,draw}} 
\begin{figure}[!ht] 
 \centering
 \subfigure[]
  {
   \begin{tikzpicture}[boximg]
    \node (img_all) {\includegraphics[width=0.42\linewidth]{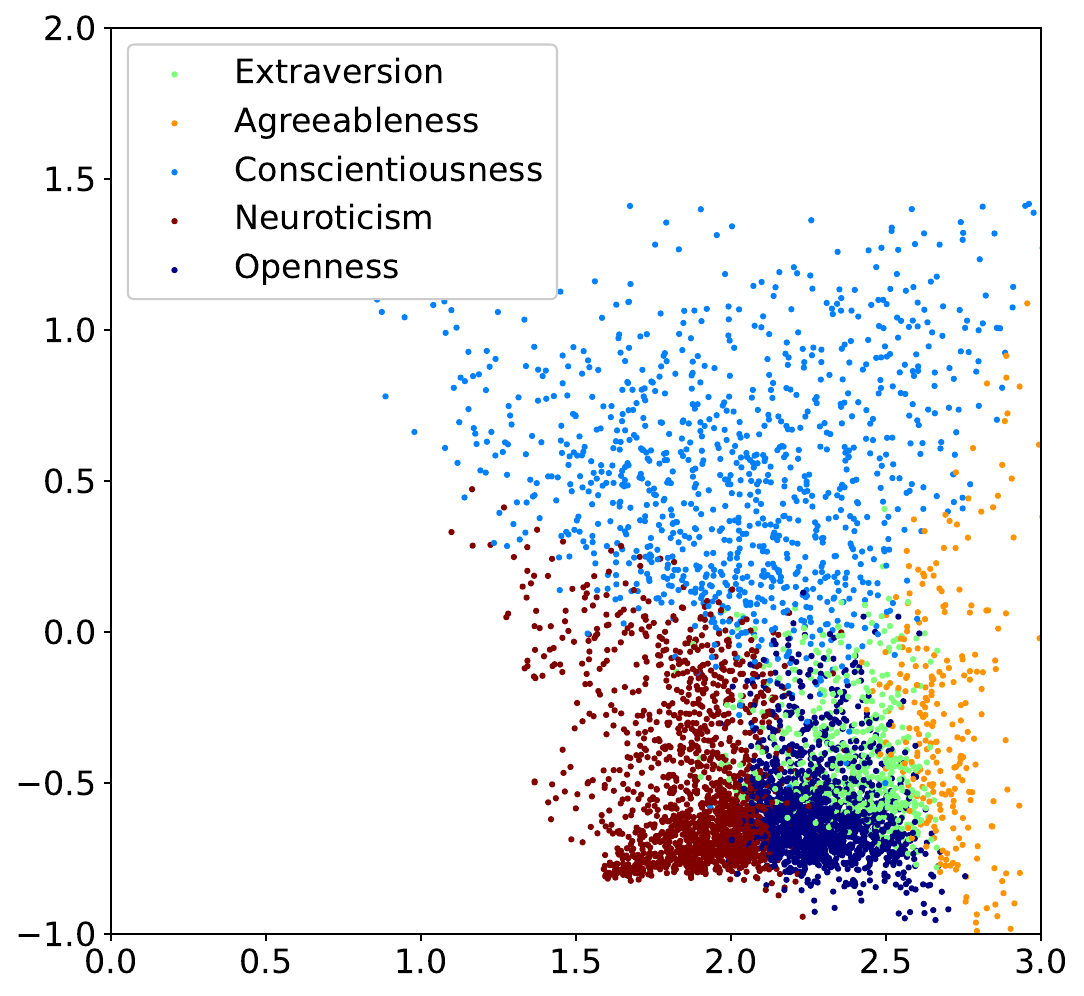}}; 
    \node[draw,minimum height=0.65cm,minimum width=0.65cm,anchor=south west] (sub1) at (0.65,-1.45) {}[blue]; 
   \end{tikzpicture} 
  }
 \subfigure[]
 {
   \begin{tikzpicture}[boximg] 
    \node (img_sub1) {\includegraphics[width=0.42\linewidth]{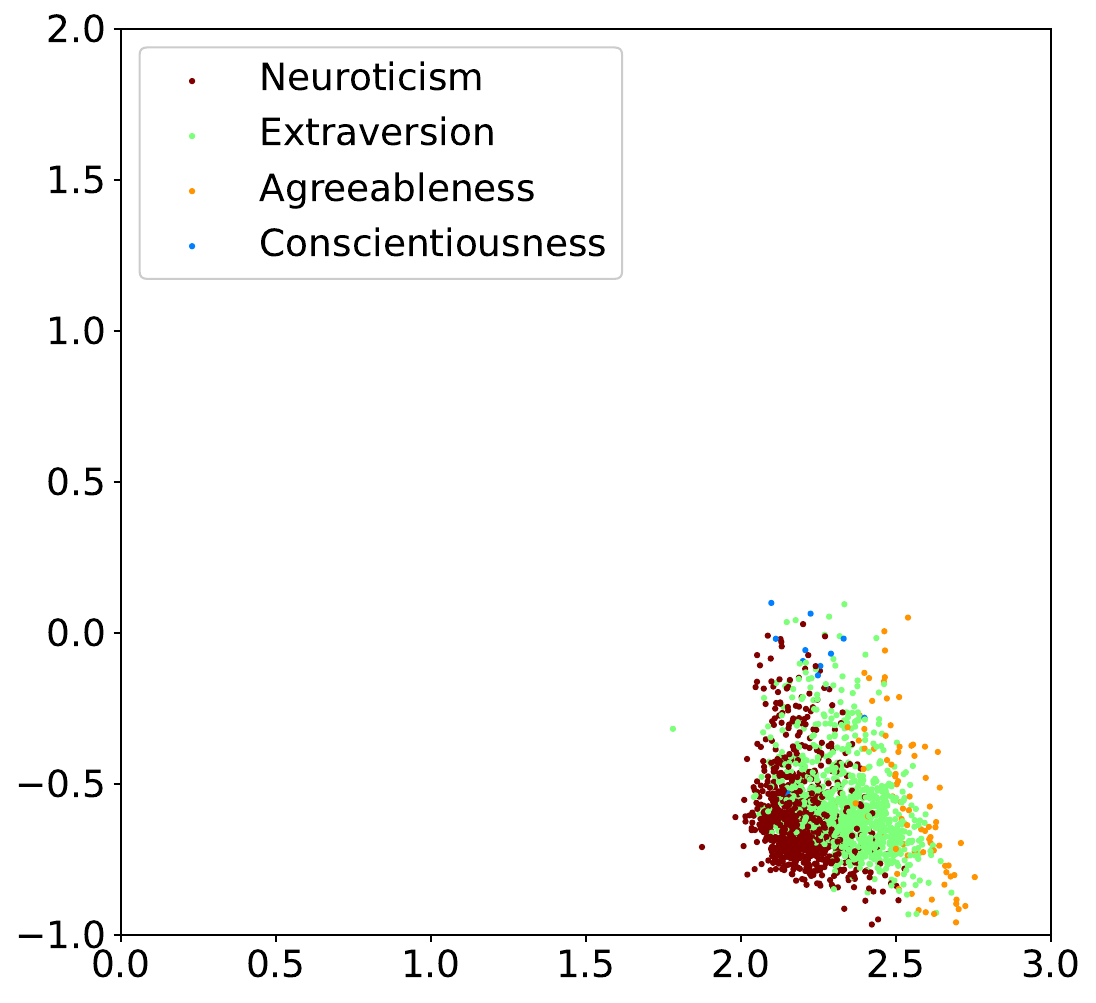}}; 
    \draw (img_sub1.south west) rectangle (img_sub1.north east)[blue]; 
    \node[draw,minimum height=1.15cm,minimum width=0.45cm,anchor=south west] (sub2) at (0.6,-1.5) {}[red]; 
   \end{tikzpicture} 
 }
 \\
 \subfigure[]
 {
   \begin{tikzpicture}[boximg] 
    \node (img_sub2) {\includegraphics[width=0.42\linewidth]{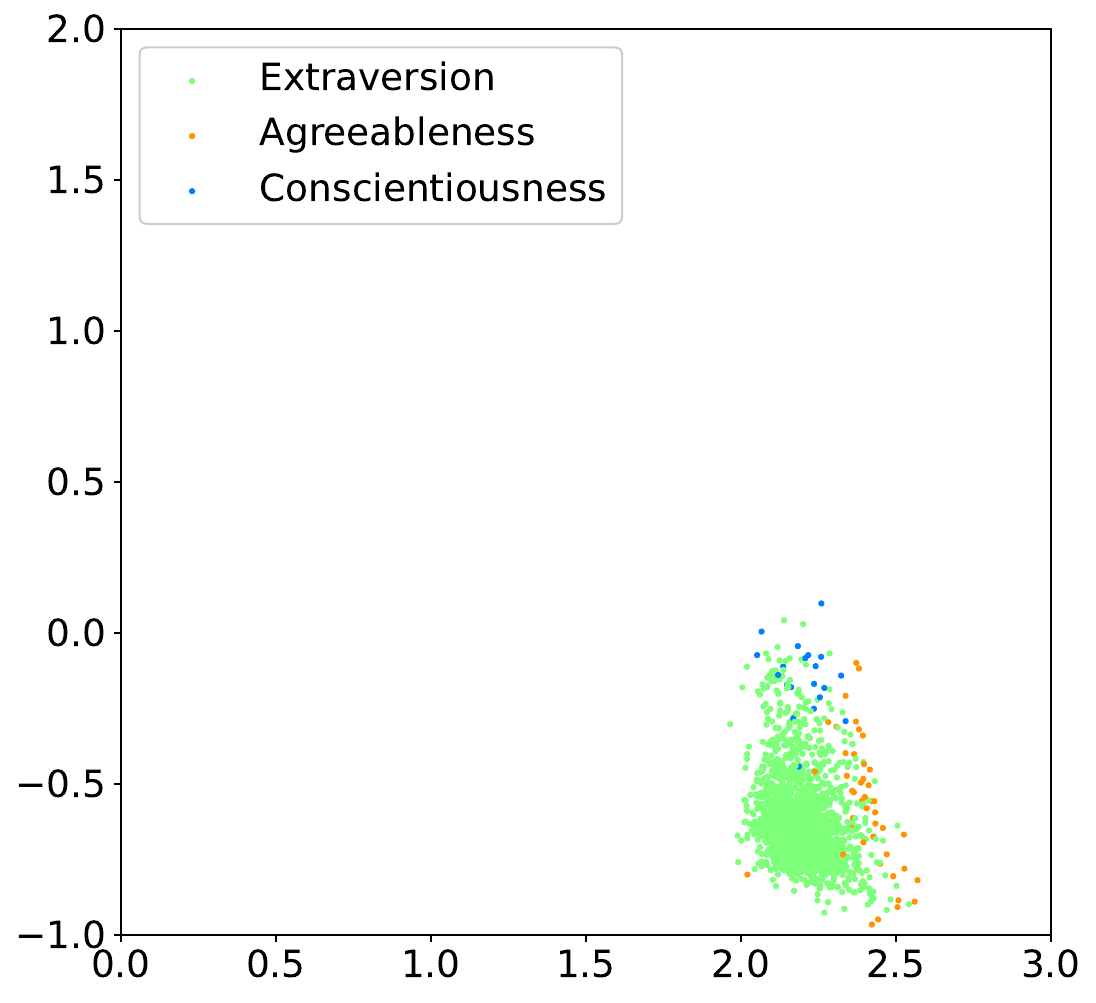}};
    \draw (img_sub2.south west) rectangle (img_sub2.north east)[red]; 
    \node[draw,minimum height=1.2cm,minimum width=0.5cm,anchor=south west] (sub3) at (0.65,-1.5) {}[green];
   \end{tikzpicture} 
 }
 \subfigure[]
 {
  \begin{tikzpicture}[boximg] 
   \node (img_sub3) {\includegraphics[width=0.42\linewidth]{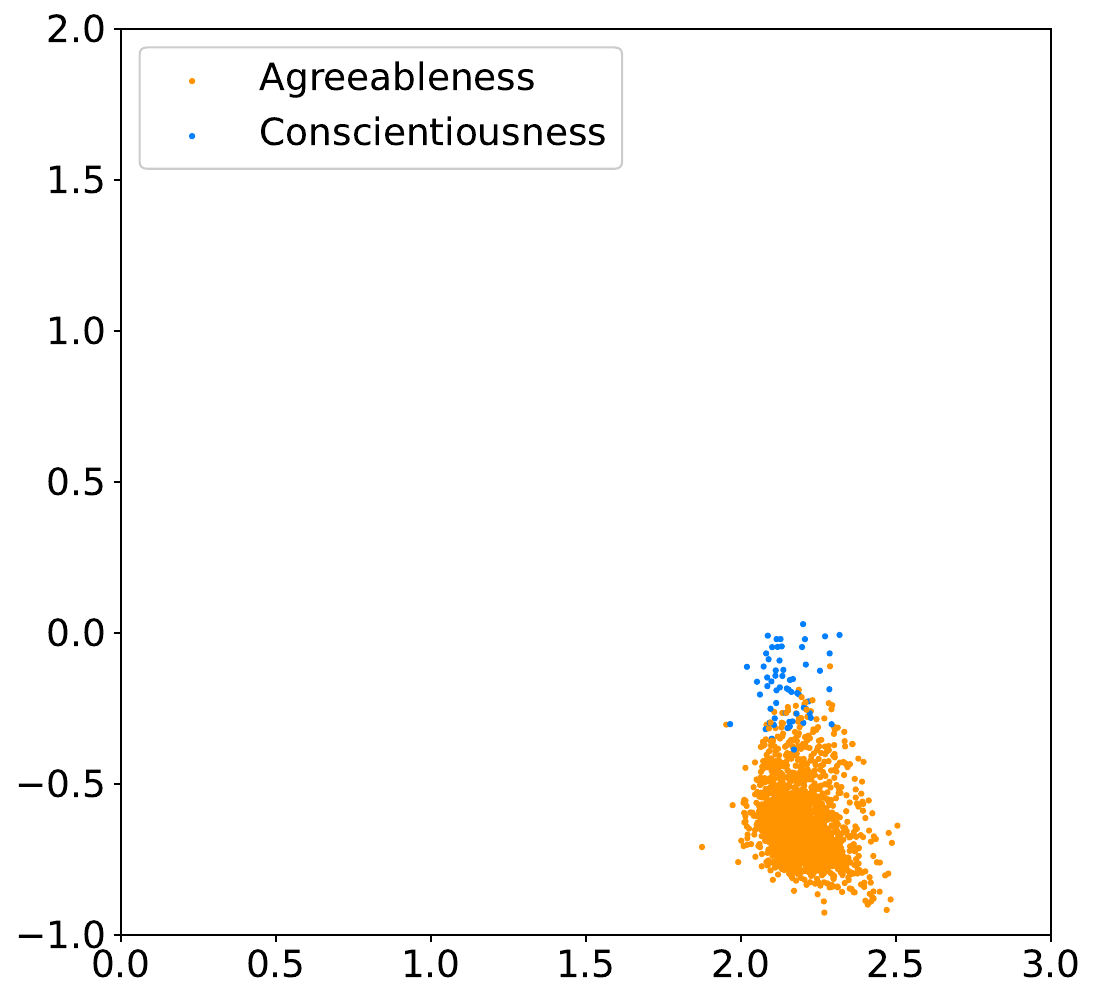}}; 
   \draw (img_sub3.south west) rectangle (img_sub3.north east)[green]; 
  \end{tikzpicture}
 }
 \subfigure
 {
 \begin{tikzpicture}[overlay,boximg] 
  \draw (sub1) -- (img_sub1)[blue]; 
  \draw (sub2) -- (img_sub2)[red];
  \draw (sub3) -- (img_sub3)[green]; 
 \end{tikzpicture} 
}
\caption{
  Hierarchical clustering of trajectory angles in 2-dimensional space. Each axis represents an angle (in radians) which describes the direction of the   trajectories in 3-dimensional space and each data point represents a trajectory. These points can be interpreted as the locations where the trajectories penetrate a sphere enclosing the state space. We show all the levels of hierarchical clustering: (a) shows the   highest level, while (b) through (d) show sub-clustering within each of the subsequent parent clusters.}
 \label{fig:hierarchy}
\end{figure}
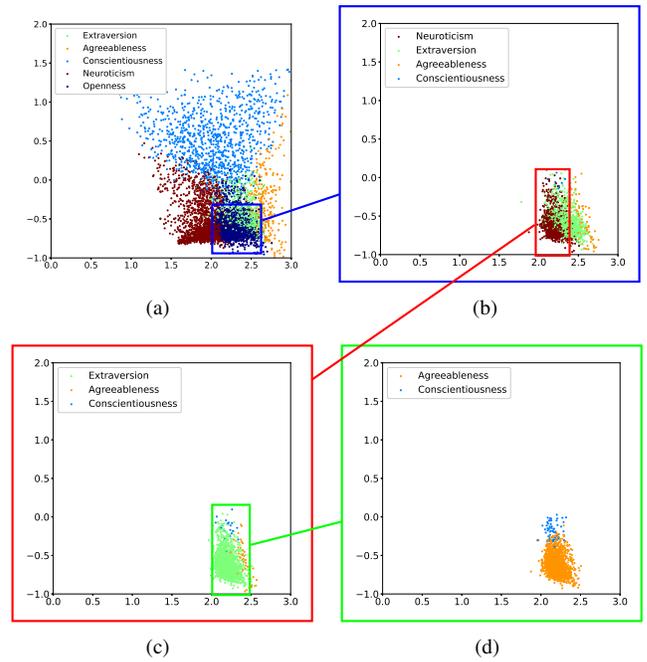

\begin{figure}[!htbp] 
 \centering
 \includegraphics[width=0.9\linewidth]{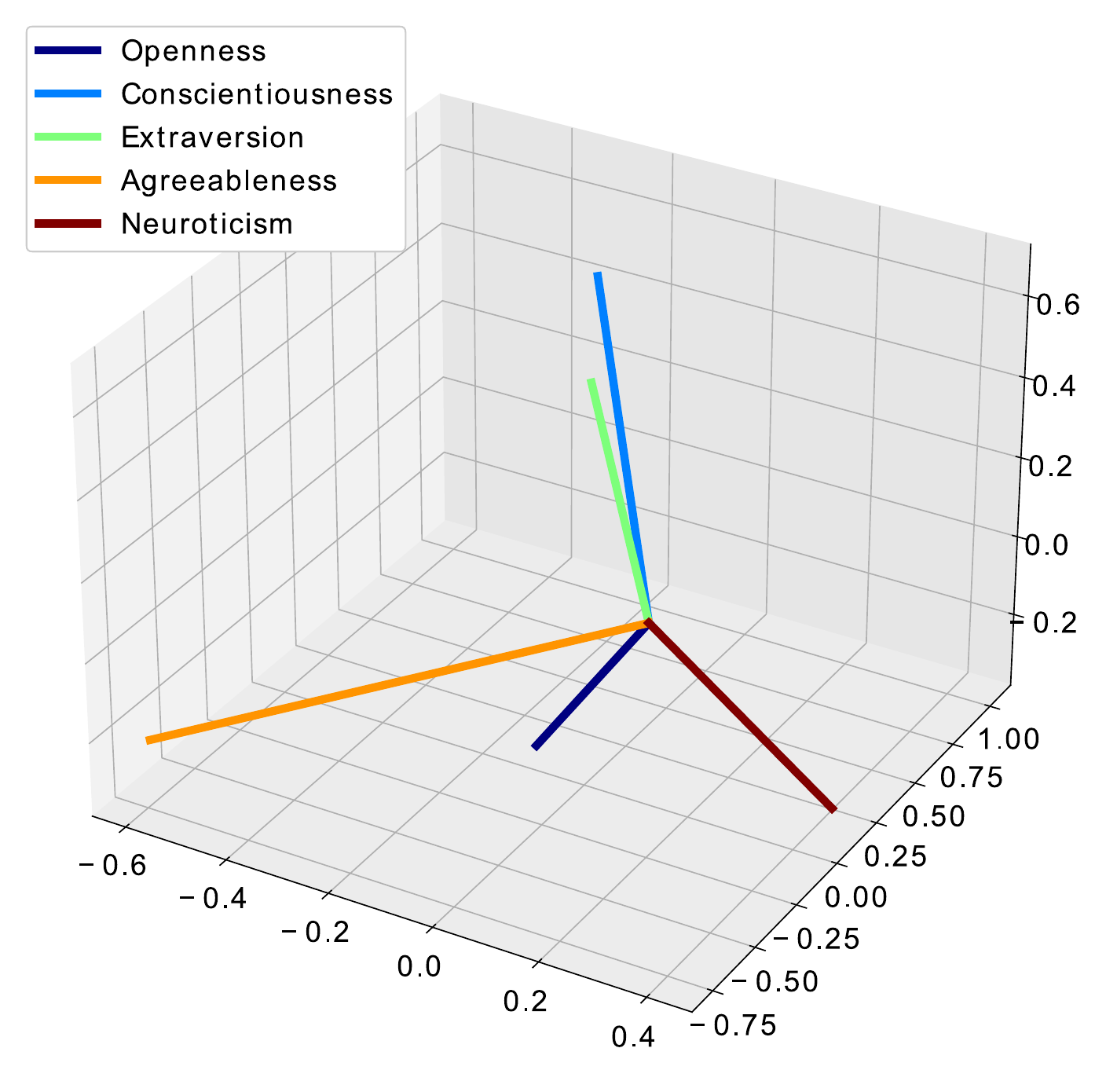} 
 \caption{The dimensions of the output space of our RNN ($O$) mapped onto the state space ($S$) as per Equation \ref{eqn:mapped_dim}. Each coloured line represents a different labelled dimension in $\mathcal{O}$, with the lengths of the lines mapped from the maximum observed values of their corresponding output dimensions (Equation \ref{eqn:D}).} 
 \label{fig:inverse_dimensions} 
\end{figure}

We observed that the directions of the trajectories were consistent with the grades of the customers' memberships in each of the five personality traits, i.e., the output data $O$ of the RNN. The greater a customer's membership in the dominant personality trait, the quicker the trajectories converged towards the corresponding hypothesised attractor. The attractors acted not only on the dominant personality trait, but also on succeedingly lesser personality traits with succeedingly lesser forces. We demonstrate this in Fig. \ref{fig:hierarchy} where the sub-clusters preserve the structure of their parent clusters: trajectories of lesser personality traits also converged to their respective attractors. Intuitively, people spend differently according to their dominant personality trait. Within a group of their peers, their lesser personality traits still differentiate them from each other. Thus the hierarchical clustering of trajectories and the labeling of the attractors is the model interpretation.  
Based on this observation and to locate and label the attractors, we mapped the dimensions of the output space $O$ onto the state space $S$ using inverse regression, as described in Section \ref{sec:inv_pred}. The resulting mapping ($\mathcal{O}$) is shown in Figure \ref{fig:inverse_dimensions} where each colored axis represents a personality dimension. These are the axes along which customers' trajectories moved in time; each time-step moved a trajectory further along these dimensions, with the direction dictated by the grades of membership in each of the output dimensions. We proved this by predicting the final location in the state space ($\mathcal{L}$) of each trajectory given the normalized grades of membership in each of the dimensions in the output space $O$.

\begin{align} 
\mathcal{L} &= \mathcal{O}^T \cdot {O^{\prime}}^T \label{eqn:predicted_location} \\
O^{\prime} &= \frac{O}{\max\limits_{1 \leq i \leq |K|}{O_{i,j}}} , \ j \in [1..P] \nonumber
\end{align}

\begin{figure}[!ht] 
 \centering
 \subfigure[]{
  \includegraphics[width=0.45\linewidth]{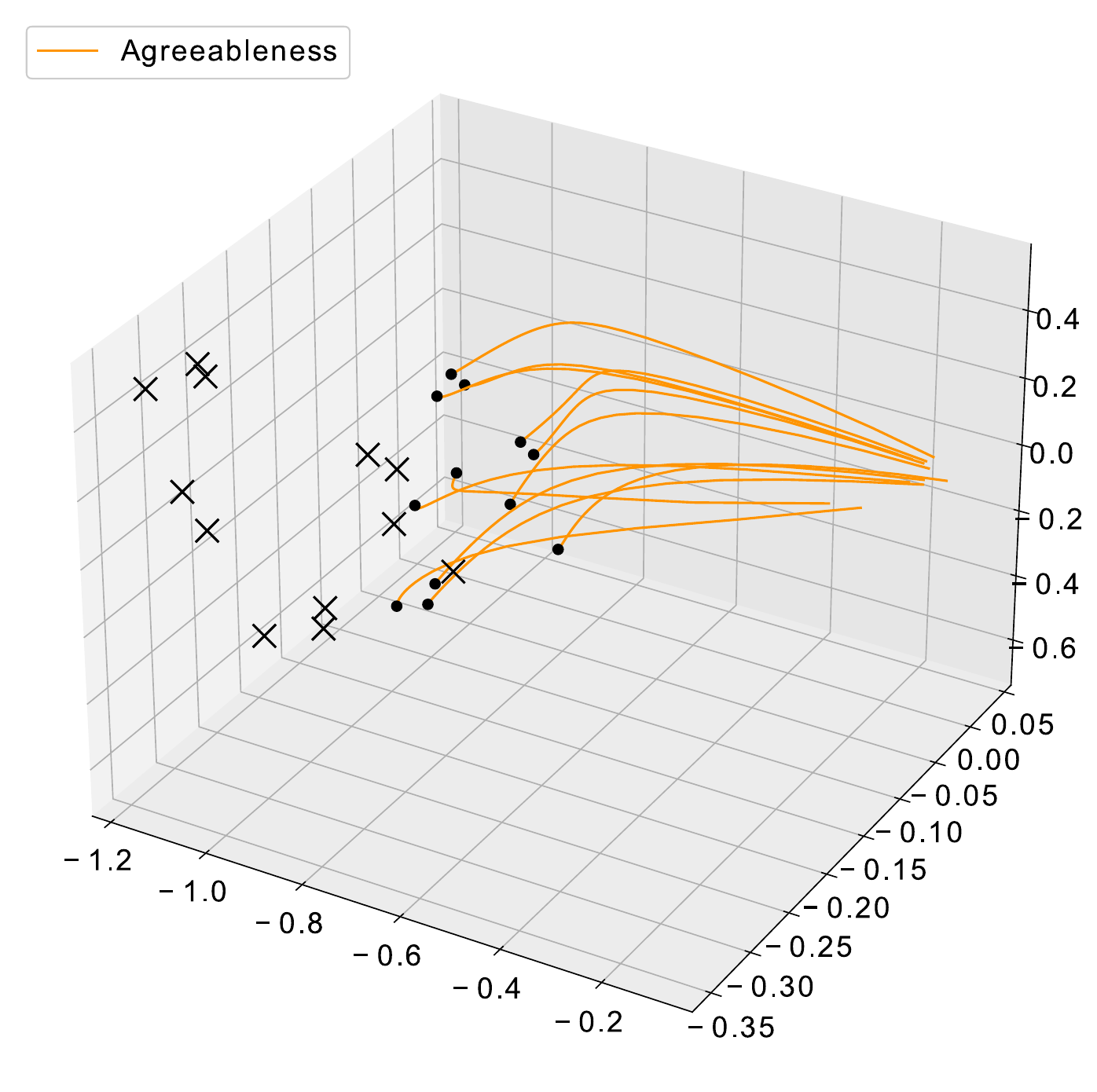} 
  \label{fig:explain_agree}
  }
 \subfigure[]{
  \includegraphics[width=0.45\linewidth]{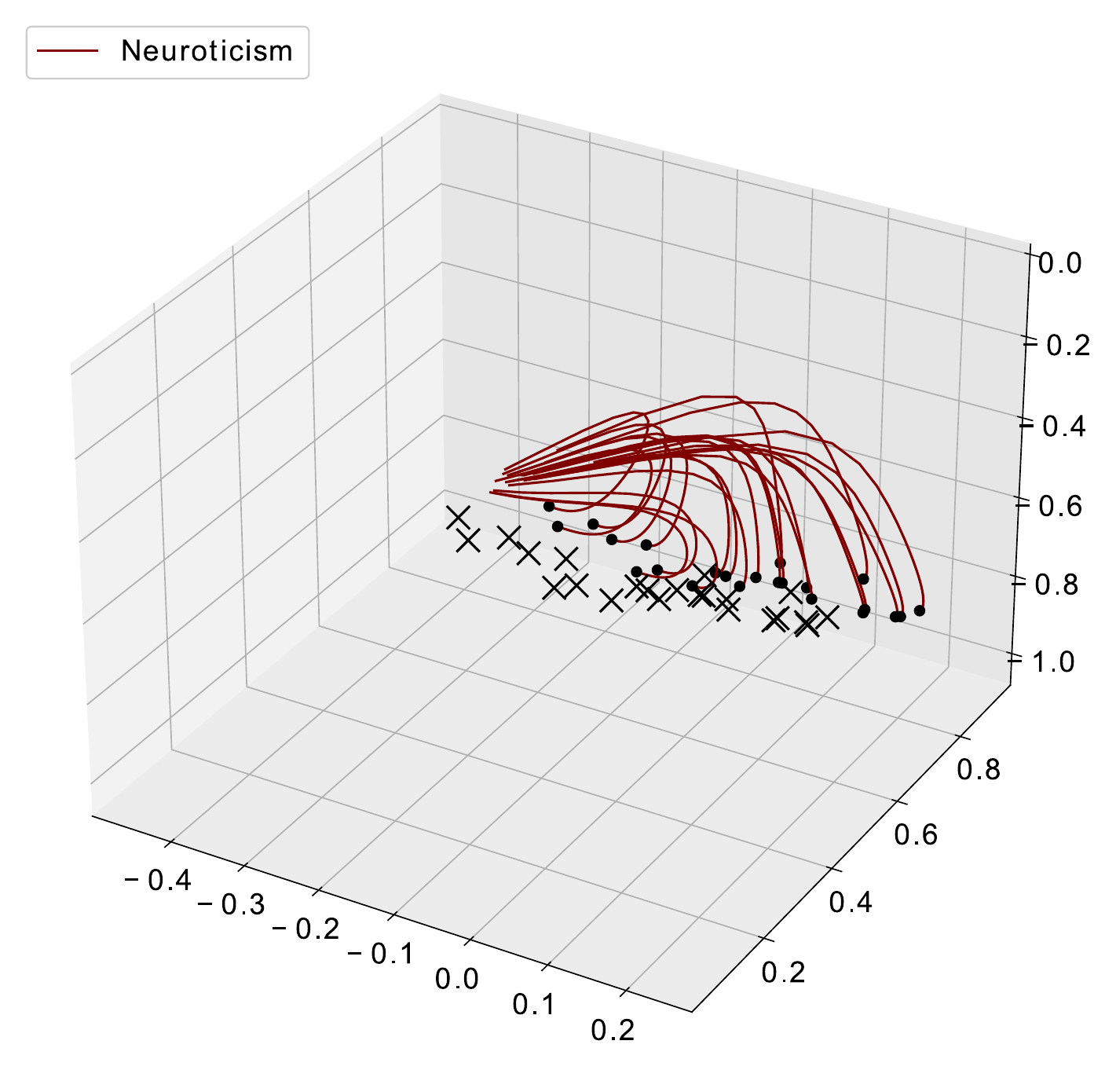} 
  \label{fig:explain_neuro}
 }
 \\
 \subfigure[]{
  \includegraphics[width=0.45\linewidth]{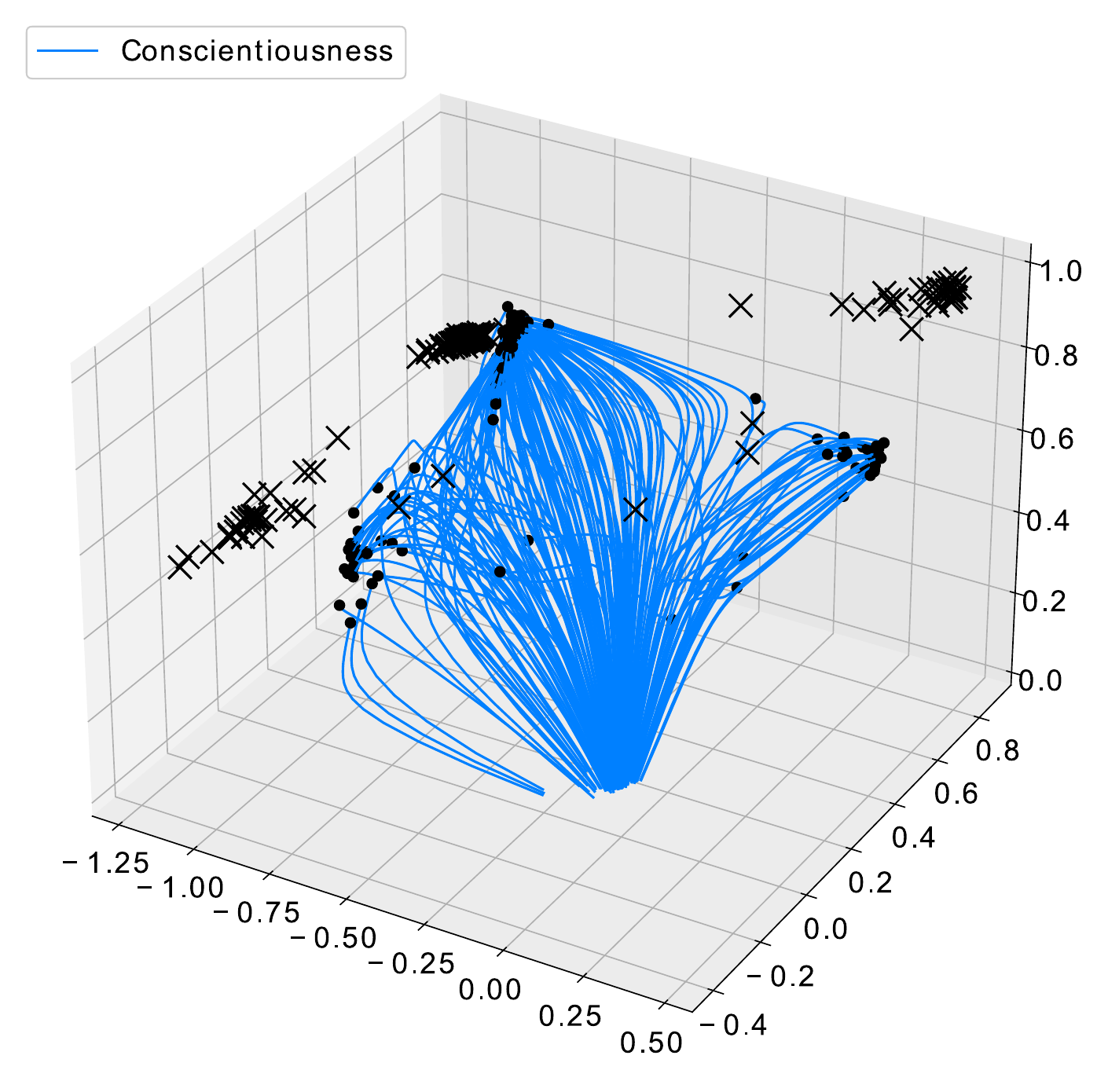} 
  \label{fig:explain_cons}
 }
 \subfigure[]{
  \includegraphics[width=0.45\linewidth]{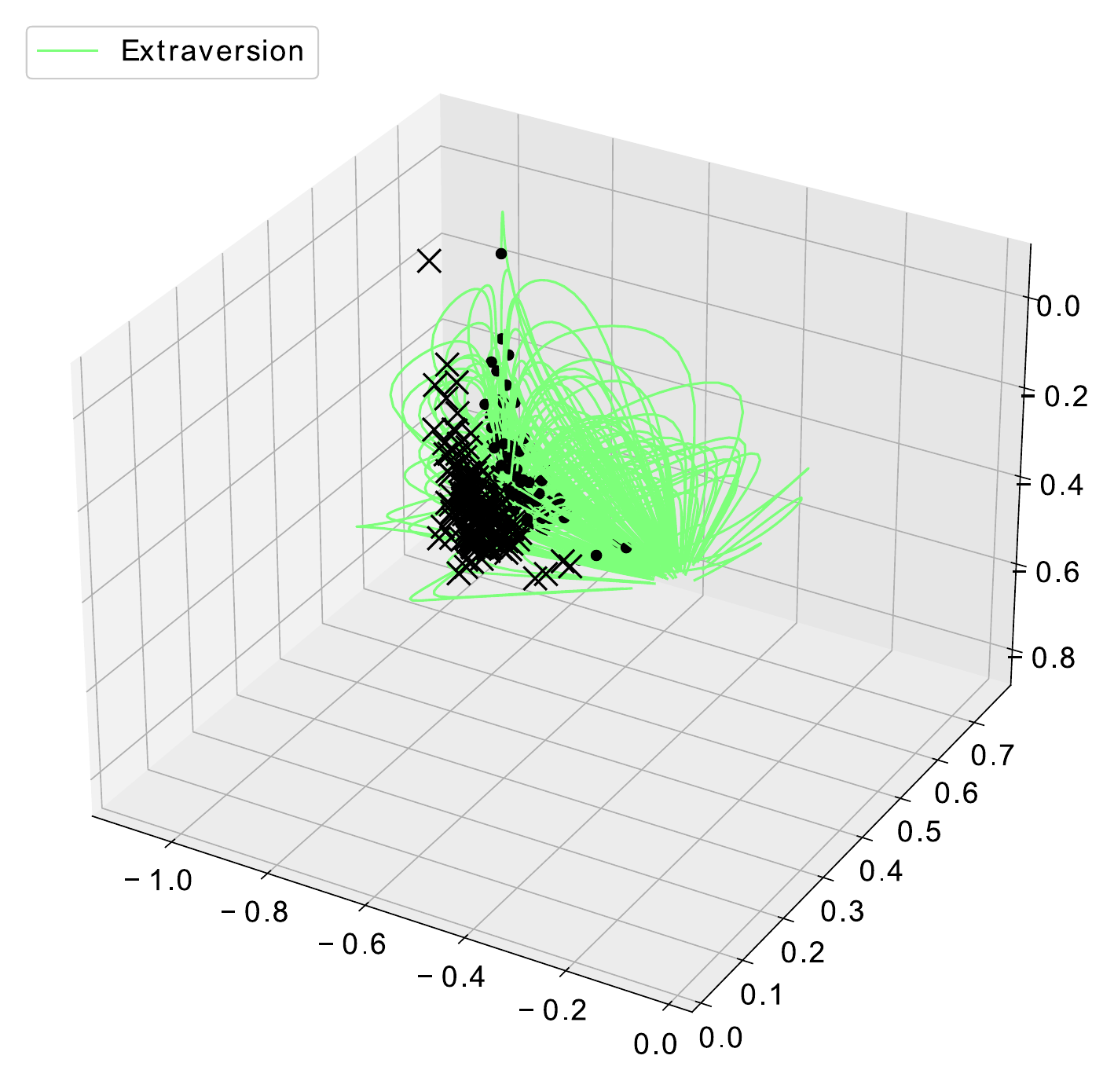} 
  \label{fig:explain_extra}
 } 
 \caption{
  Extended customer trajectories ($I'$) asymptotically converging to their predicted final locations ($\mathcal{L}$) in the state space, shown as $X$'s. Each of the sub-figures show a different cluster of customer trajectories, each having a different dominant personality trait. 
  } 
 \label{fig:projections} 
\end{figure}

\begin{figure}[!ht]
    \centering
    \includegraphics[width=0.6\linewidth]{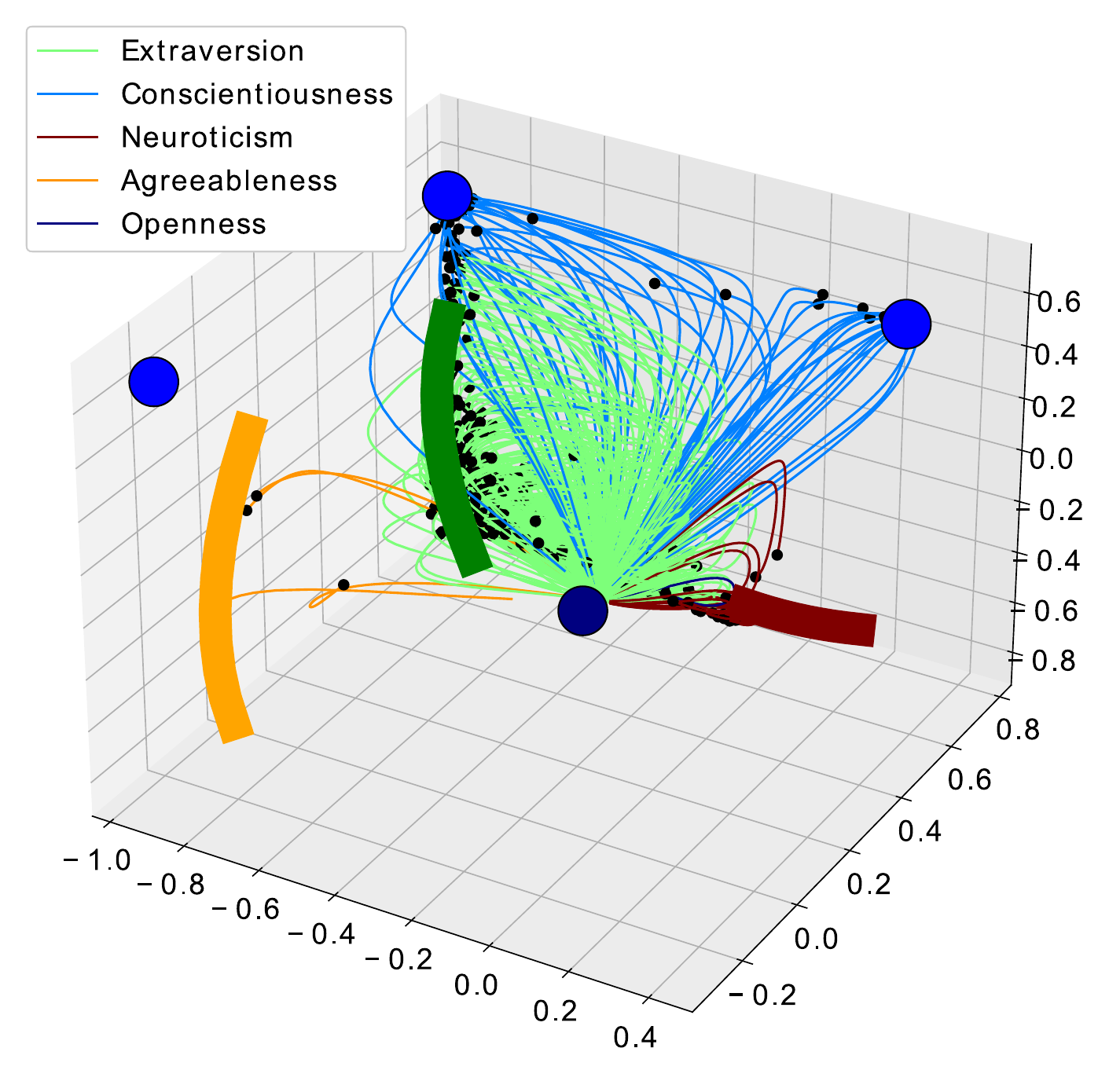}
    \caption{A subset of trajectories in $I'$ converging to their relevant attractors as determined by their dominant personality traits. The attractors are colored according to their corresponding personality traits and shown as polynomial lines (for line attractors) and circles (for point attractors). For readability, these attractors are drawn oversized as thick lines or circles.}
    \label{fig:attractors}
\end{figure}

Figure \ref{fig:projections} shows the predicted final locations ($\mathcal{L}$) of customers' \emph{extended} trajectories in the state space. We calculated these extended trajectories $I' \in [0,1]^{N \times T' \times C}$ by extending the number of time-steps to $T' = 100$, such that $I'_{n,t',c} = mean_{t \in [1,T]} (I_{n,t,c}) \ \forall \ n \in [1,N], t' \in [1,T'], c \in [1,C]$. This extension was intended to allow a larger number of time-steps such that the state space trajectories may converge to their predicted final locations $\mathcal{L}$. Note that though all trajectories \emph{asymptotically} converged towards their predicted final locations, some did not fully converge. Using the extended trajectories from Fig \ref{fig:projections}, we estimated the locations of the attractors, shown in Fig \ref{fig:attractors}. For three of the personality traits - agreeableness, extraversion and neuroticism - we observed line attractors which we located by fitting second-order polynomial functions to the final locations of the trajectories. For the remainder of the personality traits - openness and conscientiousness - we observed point attractors, with conscientiousness having three separate point attractors. We located these attractors by taking the means of the clusters of the final locations of the trajectories. Since the locations of the attractors corresponded to the predicted final locations for the trajectories $\mathcal{L}$, we could use these locations to label the attractors according to the $P$ personality dimensions in the output space $O$. The interpretation of the state space dynamics is therefore the locations and labels of the attractors based on customers' personality traits.

\newpage
\section{Conclusions and Directions for Future Work} 
The financial sector is experiencing an increased demand in the level of personalization offered to its customers, which requires more nuanced segmentation techniques than the current offerings from traditional features such as demographics. Representation learning offers such an alternative technique for fine-grained segmentation, but it is plagued by the inherent opacity introduced by deep learning;  explainability and interpretability instill understanding and are key in sensitive industries such as finance which must comply with regulations regarding the responsible use of AI. We proposed a solution for micro-segmentation of customers by extracting temporal features from the state space of a RNN which formed clusters of trajectories along the most dominant of the Big-Five personality traits. Within each such cluster we found a hierarchy of sub-clusters which corresponded to the successive levels of dominance of the personality traits. While the clusters of trajectories corresponding to the dominant personalities provide a coarse customer segmentation, the hierarchy of trajectory clusters associated with lesser personality traits offers the opportunity for micro-segmentation. 

In this study, we provided a symbolic \emph{explanation} for the RNN through a high fidelity linear regression model which answers questions such as ``Why was Customer A classified in this way?'' by referring to their historic financial transactions. 
Further, we provided an \emph{interpretation} of the feature trajectories by applying inverse regression to map the personality dimensions into the state space, which allowed us to locate and label the attractors that govern the dynamics of the state space.

In future work, we intend to use our explainable features in the development of personal financial services such as personalized savings advice, advanced product recommendations, and wealth forecasters.
\section*{Acknowledgments} 
We are grateful for fruitful discussions with Joe Gladstone on the topic of personality traits and the determination of their corresponding coefficients and with Peter Tino, Andrea Ceni, and Peter Ashwin on the topic of dynamical systems and how they apply to the evaluation of state spaces of RNNs.
\newpage
\bibliographystyle{IEEEtran}
\bibliography{references}  
\newpage
\section*{Appendix}

\vfill

\end{document}